%% file: main.tex
\documentclass[10pt, journal]{IEEEtran}

\usepackage[final]{graphicx}
\usepackage{amsmath}
\usepackage{amssymb,times}
\usepackage{cite}
\usepackage[ruled,vlined]{algorithm2e}     
\usepackage{graphicx}
\usepackage{subfigure}
\usepackage{color}
\input{com}

\usepackage{bm}

\IEEEoverridecommandlockouts

\usepackage{tabularx}

\begin{document}

\title{ A Joint Morphological Profiles and Patch Tensor Change Detection for Hyperspectral Imagery }

\author{Zengfu Hou,  Wei Li, \emph{Senior Member, IEEE}


\thanks{ Z. Hou and W. Li are with the School of Information and Electronics, Beijing Institute of Technology, Beijing 100081, China (e-mail:
zengf.hou@bit.edu.cn; liwei089@ieee.org).}

}

\maketitle
\begin{abstract}
Multi-temporal hyperspectral images can be used to detect changed information, which has gradually attracted researchers' attention. However, traditional change detection algorithms have not deeply explored the relevance of spatial and spectral changed features, which leads to low detection accuracy. To better excavate both spectral and spatial information of changed features, a joint morphology and patch-tensor change detection (JMPT) method is proposed. Initially, a patch-based tensor strategy is adopted to exploit similar property of spatial structure, where the non-overlapping local patch image is reshaped into a new tensor cube, and then three-order Tucker decompositon and image reconstruction strategies are adopted to obtain more robust multi-temporal hyperspectral datasets. Meanwhile, multiple morphological profiles including max-tree and min-tree are applied to extract different attributes of multi-temporal images. Finally, these results are fused to general a final change detection map. Experiments conducted on two real hyperspectral datasets demonstrate that the proposed detector achieves better detection performance.

\end{abstract}

\begin{IEEEkeywords}
Hyperspectral, change detection, tensor, patch strategy, morphological profiles.
\end{IEEEkeywords}

\IEEEpeerreviewmaketitle
\section{Introduction}
Change detection in hyperspectral imagery (HSI) has been a topic of long-standing interest due to its wide applications, such as missile early-warning, battlefield dynamic monitoring, environmental monitoring, land change, urban expansion, and disaster detection and evaluation, etc. Hyperspectral imaging can collect data into 3-D cubes with spatial and spectral information, where contiguous spectral information creates an opportunity for the detailed analysis and identification of the land-cover materials \cite{marinelli2019novel}.

In single-band change detection algorithms, some latent variations are hidden inside strong changes, resulting in highly mixed with each other. In contrast, multi-temporal hyperspectral images can provide continuous spectral changed information of the same imaged scenes. These outstanding advantages lead to hyperspectral change detection technology playing important roles. However, the acquisition of hyperspectral data is expensive, and these existing change detection detectors mainly developed for single-band remote sensing images are applied to hyperspectral dataset after dimensionality reduction processing, resulting in the loss of spectral information. Therefore, lower detection accuracy and higher false alarm rate are produced, which shows that they are not suitable for hyperspectral change detection. However, different from single-band change detection algorithms, there are fewer detectors developed specifically for hyperspectral change detection. Currently, hyperspectral change detection is still in development stage. Although some special detectors developed for hyperspectral data have produced good detection, they are still some difficulties and problems, which need to be further solved. 

Analyzing the existing hyperspectral change detection methods and related literature, it can be found that some algebra-based methods consider changes caused by pixel gray levels or spectral fluctuation as the main basis for evaluating material changes, such as image difference, image ratio, image regression, and absolute distance (AD) \cite{du2012fusion,bruzzone2000automatic} , etc.. These methods can be widely used for single band or multi bands change detection, which merely stack different bands information as gray levels of multiple channels. Therefore, multi-dimensional spectral information is discarded. 

The spatial coverage area of pixels may contain some different substances, where each substance has its unique spectral signal. Therefore, disturbed by the reflection of various substances, the actual spectrum obtained is a mixed signal. Subsequently, some algorithms based on spectral unmixing are developed, 
such as multitemporal spectral unmixing (MSU) \cite{liu2016unsupervised} etc.. In contrast, subspace projection transformation is considered to be another important mathematical tool for solving this kind of problem, which project original hyperspectral data into another feature subspace to increase the difference between changed and unchanged pixels, thereby marking the changed pixels. Such as conventional principal component analysis (CPCA) \cite{deng2008pca}, temporal principal component analysis (TPCA) \cite{ortiz2006change}, multivariate alteration detection (MAD) \cite{nielsen1998multivariate}, and the independent component analysis \cite{marchesi2009ica}, etc.. 

Classification-based methods \cite{bovolo2008novel,demir2011detection,ahlqvist2008extending} include post-classification and direct classification, which treat change detection as classification tasks. By classification algorithms, the postclassification method processes the images of different time series separately, and then classification results are compared and analyzed. The direct classification method stack multitemporal images together for classification task, where the same classifier is used to find changed categories.

With the development of compressed sensing and deep learning, low rank and sparse representation and deep learning-based methods are also applied to hyperspectral change detetion, such as joint sparse representation based anomalous changed detection (JCRACD) \cite{wu2018hyperspectral}, general end-to-end 2D convolutional neural network (GETNET) \cite{wang2018getnet}, and deep slow feature analysis (DSFA)\cite{du2019unsupervised}, etc.. Deep learning-based methods aim at generating a data-driven linear or nonlinear transformation to obtain advanced features of data for change detection. Therefore, the scale of training database data and the accuracy of labels determine the performance.

Recently, morphological-based method has shown some promising potential for hyperspectral processing task, where the tree theory is introduced into image processing to reflect the topological structure between objects. Considering both spatial and spectral information, Hou et al. proposed a dual-pipeline framework for hyperspectral change detection \cite{hou2021hyperspectral}, where max-tree and min-tree are used for the first time to extract morphological features. Although morphological methods shows robustness to illumination and shadow, it is mainly used to process single band data. Therefore, after dimensionality reduction of hyperspectral data, the original spectral information is lost. In traditional change detection algorithms, hyperspectral data is treated as a 2D matrix by flattening operation, which ignore the inherent structure information of hyperspectral 3D cube. Therefore, Hou et al. subsequently proposed Tucker decomposition and reconstruction detector (TDRD) \cite{hou2021three} for hyperspectral change detection. Hoverver, this detector ignores the similarity between the local data structure, which leads to the lack of full exploitation of spatial information. Taking into account the local spatial similarity of sepctrum and the global topological structure of image, a joint morphology and patch-tensor change detection (JMPT) method is proposed for hyperspectral image.

The main contributions can be summarized as follows. 1) A novel dual-pipeline framework jointing morphology and patch-tensor is proposed for hyperspectral change detection. 2) Morphological attribute profiles and tensor processing are effectively combined for hyperspectral change detection for the first time, where max-tree and min-tree in attribute profiles are stacked together to fully excavate the global topological structure of hyperspectral images. 3) Patch-based tensor decomposition and reconstruction strategy is firstly adopted to exploit nonoverlapping local spatial similarity of sepctrum structure. 4) A specially designed detector is proposed for change detection to further improve the detection accuracy.

The remainder of this paper is organized as follows. In Section \ref{sec2}, a detailed description of the proposed framework
is presented. In Section \ref{sec3}, two real datasets are utilized to verify the proposed method, and the experiment results and parameters are analyzed and discussed. The conclusion is drawn in Section \ref{sec4}

\emph{Notation}:
Vectors (matrices) are denoted by boldface lower (upper) case letters.
Superscript $(\cdot)^T$ denotes transpose, and $\mathbb{R}^{m \times n}$ is a real matrix space of dimension $m\times n$. $\Ibf_n$ stands for an identity matrix of $n\times n$, and $\mathbf{0}_{m\times n}$ represents a null matrix of dimension $m\times n$. For notational simplicity, we sometimes drop the explicit indexes in $\Ibf_n$ and $\mathbf{0}_{m\times n}$ if no confusion exists.

\section{Proposed change detection framework}
\label{sec2}

\begin{figure*}
	\centering
	\includegraphics[width=16cm]{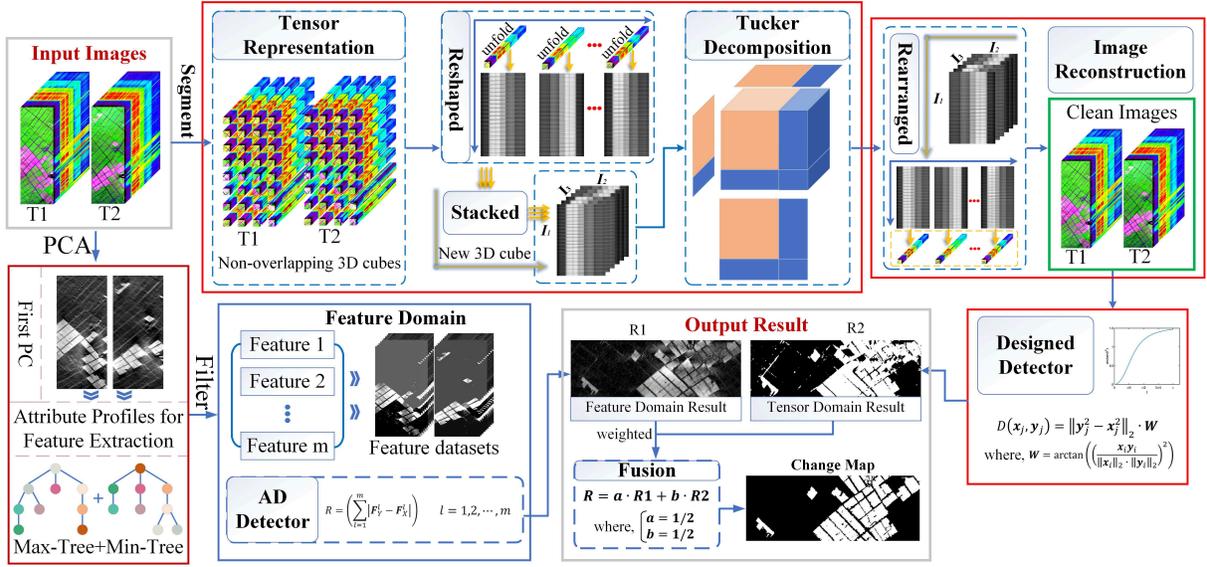}
	\caption{Framework of proposed JMPT detector for hyperspectral change detection.
		\label{Fig:Workflow}}
\end{figure*}

\subsection{Multiple Morphological Profiles (MMPs)}
  
\begin{figure}[tp]
	\subfigure[\scriptsize{}]{
		\label{Fig:subfig:Sample image}
		\includegraphics[width=2.5 cm]{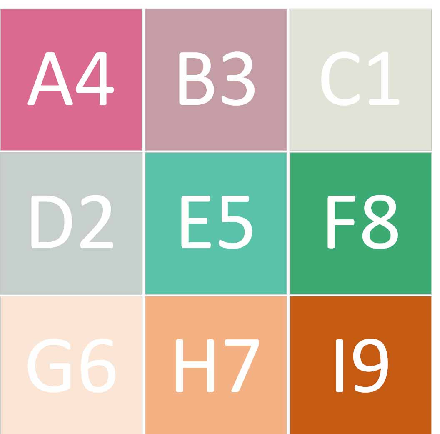}}
	\subfigure[\scriptsize{}]{
		\label{Fig:subfig:Max-tree}
		\includegraphics[width=2.5 cm]{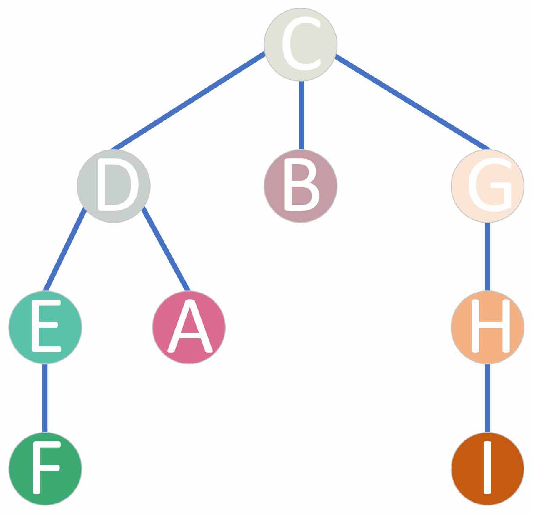}}
	\subfigure[\scriptsize{}]{
		\label{Fig:subfig:Min-tree}
		\includegraphics[width=2.5 cm]{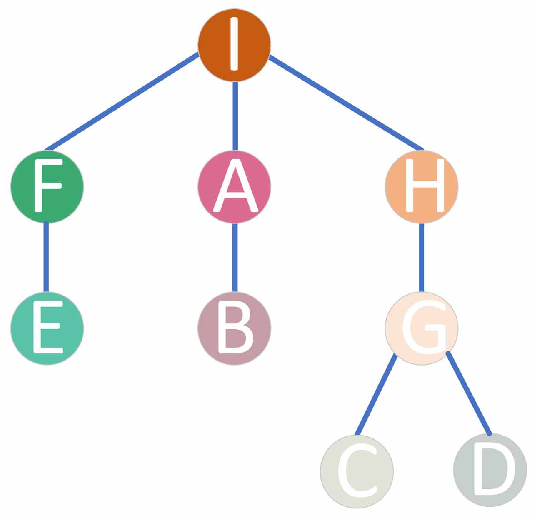}}
	
	\caption{Illustration of the max-tree and min-tree. (a) Sample image. (b) Max-tree. (c) Min-tree.}
	\label{fig:maxtree_mintree_struction}
\end{figure}

Morphological Profiles can be used to directly and accurately characterize the connected region related to objects, which have been demonstrated its utility and rigor in mathematical description. Therefore, it has attracted extensive attention in hyperspectral image processing\cite{hou2021hyperspectral,zhao2020infrared}. As mathematical morphological algorithms, max-tree and min-tree can be used to construct the tree structure of multi-temporal images, which makes it possible to exploit both contexture and spatial information.

For this purpose, attribute profiles (APs) of max-tree and min-tree corresponding to image are used to constructed a morhological feature space for exploring changed objects. The max-tree/min-tree processing mainly consists of three steps: (1) tree construction, (2) filtering/pruning, and (3) image reconstruction. More details can be found in \cite{hou2021hyperspectral}. However, morphological methods \cite{dalla2010morphological} are mainly used to process natural images. If each band of hyperspectral image is processed by morphological filtering, it inevitably causes a very high-dimensional feature space \cite{hughes1968mean}. For avoid this, principal component analysis (PCA) is firstly adopted to reduce the dimensionality of the original hyperspectral image, where the first principal component is selected for morphological feature extraction. 
 
Max-tree and min-tree are structured representations of connected components with different level sets. In the max-tree, its value gradually increases from root node to leaf node, that is, the value of leaf node is greater than that of root node. However, in the min-tree, the closer to the leaf node, the smaller its value. In Fig. \ref{Fig:subfig:Sample image}, a sample image is displayed to understand the difference between max-tree and min-tree, in which various colors represent different pixel values. The darker the color is, the greater the value is. As shown in Fig. \ref{Fig:subfig:Max-tree} and Fig. \ref{Fig:subfig:Min-tree}, the image-level sets can be represented by max-tree and min-tree, respectively. From Fig. \ref{Fig:subfig:Max-tree}, the value of root node is 1, and leaf nodes are 8 and 9, respectively. That is, for max-tree, the region of root node (C) has the minimum pixel value, and leaf nodes (F, I) correspond to the maximum pixel values. In contrast, in the min-tree shown in Fig. \ref{Fig:subfig:Min-tree}, the region of root node (I) has the maximum pixel value, and leaf nodes (C, D) correspond to the minimum pixel values. Therefore, the structure of max-tree and min-tree is different.

\subsection{Pruning/Filtering Process}   
After max-tree and min-tree being constructed, a pruning strategy/attribute filtering \cite{ghamisi2016extinction,dalla2010classification} is taken to keep branches of leaf nodes that meet requirements, and to remove branches of leaf nodes that do not meet the requirements. Pruning strategy/attribute filtering is powerful tools that can be used to measure the existence of regional extreme values. In APs processing, attributes are usally divided into increasing and nonincreasing ones. Depending on whether the attributes are increasing or not, corresponding tree pruning strategies or tree nonpruning strategies are chosen. If a node is filtered by a pruning strategy, then all its descendants are also filtered. In the pruning/filtering processing, removal or preservation of the node is determined as follows, 
\begin{equation}\label{eq1}
	\begin{array}{*{20}{l}}
		N_i \quad is 
		\begin{cases}
			removed,   \quad if \quad R_{N_i} <  \eta\\
			preserved, \quad if \quad R_{N_i} \ge \eta\\
		\end{cases}.
	\end{array}
\end{equation}
First, $R_{N_i}$ corresponding result value is calculated according to the specific attribute. Second, $\eta$ being a threshold determines whether the node $N_i$ should be removed or not. If $R_{N_i}$ is below $\eta$, then the corresponding node is removed, vice versa.

\begin{figure}
	\centering
	\includegraphics[width=8cm]{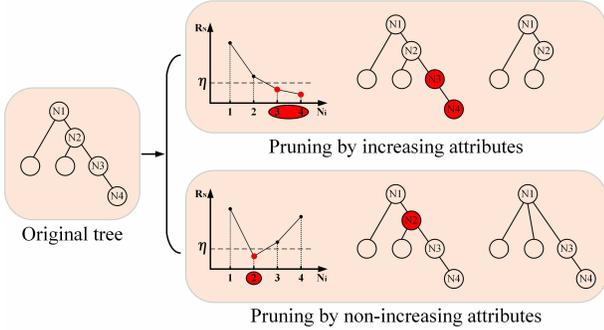}
	\caption{Compare increasing attributes and nonincreasing attributes (red circles represent removed nodes, others represent preserved nodes).}
	\label{Fig:PurningStrateg}
\end{figure}   
   
In Fig. \ref{Fig:PurningStrateg}, pruning/filtering strategies by increasing and non-increasing attributes are illustrated, where an original tree and two types results of different attributes for one local maximum are used for example. In this coordinate system, curves of increasing and nonincreasing attributes are provided, where abscissa denotes node $N_i$ ($i$ represents $i$th node), and ordinate represents corresponding results $R_{N_i} $. When root node $N_1$ is chosen, path to leaf node $N_4$ is $N_1-$to$-N_2-$to$-N_3-$to$-N_4$. For increasing attributes, the operation of tree pruning strategy is straightforward when increasing attributes are employed. Since $R_{N_3}$ and $R_{N_4}$ are less than $\eta$, when $N_3$ is removed, $N_4$ which is descendant of $N_3$ is also removed. However, in nonincreasing attributes, only $R_{N_2}$ is less than $\eta$, when $N_2$ is removed, $N_3$ and $N_4$ which is the descendants of $N_2$ are preserved. Therefore, it can be concluded that all the descendants are removed when increasing attributes are used, but descendants may be preserved when the nonincreasing attributes is employed.

The attribute values of pixels or connected regions are calculated to represent attributes of nodes in tree. In this work, five APs \cite{zhao2020infrared} are mainly used to extract the morphological features, including four increasing
attributes, i.e., area attribute, height attribute, volume attribute, the diagonal of bounding box attribute, and one nonincreasing attributes, i.e., standard deviation attribute. 

These morphological attributes are calculated differently, so the mathematical meanings they represented are also different and unique. In these attributes, area attribute $A_N$ is a scale attribute, which is to calculate the number of pixels in connected region.
\begin{equation}\label{eq2}
	\begin{array}{*{20}{l}}
		$$A_N=\{\#p|p \in N\}$$,
	\end{array}
\end{equation}
where $N$ is a connected region, which represents the node of max-tree/min-tree. $p$ represents the pixels that belong to $N$, and $\#p$ is the number of pixels. The height attribute $H_N$ is a contrast attribute, which is to calculate difference between pixels in connected region and local pixel, as follow,
\begin{equation}\label{eq3}
	\begin{array}{*{20}{l}}
		$$H_N=max_{p \in N} f(p)-min_{p \in N} f(p)$$,
	\end{array}
\end{equation}
where $f(p)$ is the gray value of pixels. The volume attribute $V_N$ is both a contrast attribute and a scale attribute, and the diagonal of the bounding box attribute $D_N$ is a combination of shape and scale attribute. These attributes are to calculate the variance of connected region, which are calculated as \cite{zhao2020infrared},
\begin{equation}\label{eq4}
	\begin{array}{*{20}{l}}
		$$ V_N=\sum_{p \in N} (max_{p \in N} g(p)-g(p)) $$,
	\end{array}
\end{equation}
\begin{equation}\label{eq5}
	\begin{array}{*{20}{l}}
		$$ D_N=\sqrt{(x_{p,max}-x_{p,min})^2+(y_{p,max}-y_{p,min})^2} $$,
	\end{array}
\end{equation}
where $g=\pm f$ is determined according to the direction, and $x_{p,max}$, $x_{p,min}$, $y_{p,max}$, $y_{p,min}$ are the extremum on the abscissa and ordinate in connected region, respectively. $(x_{p,max}-x_{p,min})$ and $(y_{p,max}-y_{p,min})$ are the maximum height and width of connected region, respectively. Different from these attributes mentioned above, standard deviation is a contrast attribute, which is calculated as,
\begin{equation}\label{eq6}
	\begin{array}{*{20}{l}}
		$$ Std(N)=\sqrt{\frac{1}{area(N)}\sum_{\forall p \in N}(f(p)-K_{g}(\lambda))} $$,
	\end{array}
\end{equation}
where $Std$ is an abbreviation for standard deviation. $K_{g}$ is the average intensity value of the pixel in the connected area, which is given by,
\begin{equation}\label{eq7}
	\begin{array}{*{20}{l}}
		$$ K_{g}(\lambda)=\frac{1}{area(N)}\sum_{\forall p \in N}f(p) $$.
	\end{array}
\end{equation}

After pruning of max-tree/min-tree by attribute values, the tree is reconstructed into a new feature image, where useful feature information is retained, while useless feature information is deleted. Therefore, after nodes on many branches being removed, connected regions corresponding to each node in this tree are to changed. Finally, this feature images reconstructed by max-tree and the min-tree are stacked together for processing. More details can be found in \cite{hou2021hyperspectral}.

\subsection{Patch-Tensor Process}
In morphological features extraction process, the spatial information of hyperspectral image is fully utilized, but the contribution of spectral information to change detection is completely ignored. However, in traditional hyperspectral change detection algorithms, spectral fluctuation of unchanged pixels caused by various factors has become one of the main challenges. Therefore, solely using spectral information cannot effectively judge changed objects in image scenarios. Tensor can effectively represent intrinsically spectral structure information of hyperspectral dataset, so more of its advantages are reflected in the change detection of various scenarios. In this work, consindering the similarity between the local data structure, patch-tensor strategy \cite{huang2017hyperspectral} is adopted to incorporate the nonlocal similar property to exploit spectral structural information.

Hyperspectral image is represented as a three-order tensor $\mathcal{Y} \in R^ {H \times W \times D}$, where $H$, $W$, $D$ represent the image rows, columns and bands, which correspond to the mode-1, model-2 and mode-3, respectively. First, tensor $\mathcal{Y}$ is divided into non-overlapping 3D cubes,

\begin{equation}\label{eq8}
	\begin{array}{*{20}{l}}
		$$ \mathcal{P}_{i,j,:} = Patch(\mathcal{Y}) $$.
	\end{array}
\end{equation}
where $Patch(\cdot)$ represents tensor division operation. $\{ \mathcal{P}_{i,j,:}\}_{1 \leq i \leq m, 1 \leq j \leq n , :} \in  R^ {w \times w \times D}$ is the patch tensor, in which $m = \lfloor H/w \rfloor$, $n = \lfloor W/w \rfloor$, $w$ is the patch size, and $\lfloor \cdot \rfloor$ represents the floor operation. Then, these 3D cubes are unfolded to form 2D matrices according to mode-3, and these matrices are stacked into new tensors dataset, as follow,
\begin{equation}\label{eq9}
	\begin{array}{*{20}{l}}
		$$ \mathcal{X}_{:,:,k} = stack_k(unfold_3(\mathcal{P}_{k})), 
		\\   k=1,2, \cdots, m \times n $$,
	\end{array}
\end{equation}
where $\mathcal{P}_{k}$ represents the $k$-th patch tensor, $unfold_3(\cdot)$ is the unfold operation of tensor according to mode-3, and $stack_k(\cdot)$ is tensor stacking operation, where a 2D matrix is stacked into $k$-th slice of tensor. $\mathcal{X} \in R^ {I_1 \times I_2 \times I_3}$ is the new nonlocal similar 3D cube tensor, in which $I_1 = w \times w$, $I_2 = D$, and $I_3 = m \times n$. However, in $\mathcal{X}$, abundant spectral structure information and noise are contained, which will amplify the spectral difference of unchanged objects in various scenarios. In order to obtain more pure dataset, tucker decomposition \cite{zhang2016tensor} is employed to reconstruct multi-temporal dataset. Tucker decomposition is regarded to be a higher order extension of singular value decomposition, which factorises a tensor into a core tensor and some factor matrices. Therefore, the tensor $\mathcal{X}$ is approximately denoted by,
\begin{equation}\label{eq10}
	\begin{array}{l}
		$$ \mathcal{X} \approx \mathcal{G} \times_1 \bm{U} \times_2 \bm{V} \times_3 \bm{W} $$, 
	\end{array}
\end{equation}
where $\mathcal{G} \in R^ {I_1 \times I_2 \times I_3}$ is core tensor, which is essentially a compressed version of the data, and its elements denote the level of interaction between distinct components. $\bm{U}\in R^ {I_1 \times I_1}$, $\bm{V}\in R^ {I_2 \times I_2}$, and $\bm{W}\in R^ {I_3 \times I_3}$ are three factor matrices, which can be consindered to be analogous to singular values, and be regarded as the principal components in each mode. The optimization problem is stated as,
\begin{equation}\label{eq11}
	\begin{array}{l}
		argmin_{ \hspace{1mm} \mathcal{G},\bm{U},\bm{V},\bm{W}} \Vert \mathcal{X} - \mathcal{G} \times_1 \bm{U} \times_2 \bm{V} \times_3 \bm{W} \Vert_F^{2} \\
		s.t.\hspace{2mm}
		\left\{
		\begin{aligned}
			\mathcal{G} \in R^ {I_1 \times I_2 \times I_3}\\
			\bm{U}\in R^ {I_1 \times I_1}, \bm{V}\in R^ {I_2 \times I_2}, \bm{W}\in R^ {I_3 \times I_3}\\
			\bm{U}^T\bm{U}=\bm{I_1},\bm{V}^T\bm{V}=\bm{I_2},\bm{W}^T\bm{W}=\bm{I_3}.
		\end{aligned}
		\right.
	\end{array}
\end{equation}

The factor matrices $\bm{U}$, $\bm{V}$, and $\bm{W}$ are orthogonal to each other, and the core tensor $\mathcal{G}$ is obtained by,
\begin{equation}\label{eq12}
	\begin{array}{l}
		$$ \mathcal{G} \approx \mathcal{X} \times_1 \bm{U} \times_2 \bm{V} \times_3 \bm{W} $$. 
	\end{array}
\end{equation}
Therefore, the optimization problem in Eq. \ref{eq2} is converted as,
\begin{equation}\label{eq13}
	\begin{array}{l}
		argmax_{ \hspace{1mm} \bm{U},\bm{V},\bm{W}} \Vert \mathcal{X} \times_1 \bm{U} \times_2 \bm{V} \times_3 \bm{W} \Vert_F^{2} \\
		s.t.\hspace{2mm}
		\left\{
		\begin{aligned}
			\mathcal{G} \in R^ {I_1 \times I_2 \times I_3}\\
			\bm{U}\in R^ {I_1 \times I_1}, \bm{V}\in R^ {I_2 \times I_2}, \bm{W}\in R^ {I_3 \times I_3}\\
			\bm{U}^T\bm{U}=\bm{I_1},\bm{V}^T\bm{V}=\bm{I_2},\bm{W}^T\bm{W}=\bm{I_3}.
		\end{aligned}
		\right.
	\end{array}
\end{equation}

Eq. \ref{eq13} is usually solved by the alternating least squares (ALS) algorithm, where any factor matrix can be obtained by eigenvalue decomposition when the other two matrices are fixed. Generally, tucker decomposition can be regarded as high-order principal component analysis (PCA), which provides simple compression to preserve $r$ principal component. But different from PCA, tucker decomposition can effectively retain the spectral structure information. Moreover, the reconsturcted tensor $\mathcal{\tilde{X}}$  has the same spectral dimension as the $\mathcal{X}$, which is obtained by,
\begin{equation}\label{eq14}
	\begin{array}{l}
		$$ \mathcal{\tilde{X}} \approx \mathcal{G}_r \times_1 \bm{U_r} \times_2 \bm{V_r} \times_3 \bm{W_r} $$, 
	\end{array}
\end{equation}
where $\mathcal{G}_r=\mathcal{G}(1:r,1:r,1:r)$, $\bm{U_r}=\bm{U}(1:r,1:r)$, $\bm{V_r}=\bm{V}(1:r,1:r)$, and $\bm{W_r}=\bm{W}(1:r,1:r)$. $r$ is the number of principal component of differnt factor matrix, which can be determined by experience. In this paper, the size of patch is smaller. In order to reduce the interference of parameters and make prgoram more automated, thereby $r$ is set to $min(I_1,I_2,I_3)$.

After obtaining the reconstructed tensor $\mathcal{\tilde{X}} \in R^ {I_1 \times I_2 \times I_3}$, each band of which is reconstructed into a 3D matrix patch $\{ \mathcal{\tilde{P}}_{i,j,:}\}_{1 \leq i \leq m, 1 \leq j \leq n, : } \in  R^ {w \times w \times D}$. Then, these patchs $\mathcal{\tilde{P}}_{i,j,:}$ are rearranged to obtain a clean dataset $\mathcal{\tilde{Y}}$ without noise. It should be emphasized that when $w*m < H$ or $w*n < W$, the size of dataset $\mathcal{\tilde{Y}}$ is smaller than original dataset $\mathcal{Y}$. To ensure the integrity of dataset, the boundary area is filled with original dataset. Therefore, the final result $\mathcal{\tilde{Y}} \in R^ {H \times W \times D}$ has the same size as original $\mathcal{Y}$. After the same processing of bi-temporal datatsets are completed, the reconstructed new datasets $\mathcal{\tilde{Y}}_{T_1}$ and $\mathcal{\tilde{Y}}_{T_2}$ corresponding to the scene at time $T_1$ and $T_2$, respectively, are carried out to change detection.

\subsection{Change Detection for Multi Domain Features}

After morphological processing, feature maps obtained by different attributes are stacked together to form a feature image. Different from traditional spectral vectors, these pixel vectors composed of spatial feature maps cannot be directly used for object recognition because spectral information details are lost with the decrease in feature dimensions. Therefore, the performance using spectral-based change detectors for detection is limited. Consindering the rubustness of these morphological features to illumination and shadow, changed backgroud objects can be identified by the magnitude of multi-dimensional difference feature maps. Therefore, AD method is employed reformulated for these feature images. The advantage of AD is simple and intuitive, and the result is easy to be interpreted, which is expressed as,
\begin{equation}\label{eq15}
	\begin{array}{*{20}{l}}
		$$ R_1=\sum_{l=1}^{m}\vert \Fbf_{T_1}^l - \Fbf_{T_2}^l \vert, \quad l=1,2,\dots,m$$,
	\end{array}
\end{equation}
where $m$ is the number of feature maps, and $\Fbf_{T_1}^l$ and $\Fbf_{T_2}^l$ represent the $l$-th feature at time $T_1$ and $T_2$, respectively. 

For the bi-temporal dataset after tensor processing,  spectral fluctuation in unchanged pixel pairs is narrowed down, which is more conducive to the fully mining of neighborhood information of testing pixel. However, various detection algorithms are developed for different applications, which leads to some changes are difficult to be reflected in traditional detectors. Therefore, it is necessary to develop a new detector for the characteristics of tensor data to amplify spectral difference of changed spectral signals and suppress background of unchanged spectral signals. Simultaneously, considering the similarity between local background pixels and testing pixels, especially eight nearest pixels around the testing pixel, which are very similar spectrally and spatially, a newly designed detector is proposed for change detection to further
improve the detection accuracy, which can be expressed as,

\begin{equation}\label{eq16}
	\begin{array}{*{20}{l}}
		$$ r_i=\vert\vert\sum_{j=1}^{8} ( \ybf^2_{j} - \xbf^2_{j})\vert\vert_2 \cdot \Wbf $$,
	\end{array}
\end{equation}
where $\xbf_j$ and $\ybf_j$ denote the $j$-th background pixels in the eight neighboring pixels corresponding to the scene at time $T_1$ and $T_2$, respectively. $\vert\vert \cdot \vert\vert_2$ is the 2-norm of vector, and $W$ represent a revised spectral angle weighted, which can be expressed as,
\begin{equation}\label{eq17}
	\begin{array}{*{20}{l}}
		$$ \Wbf = \arctan((\frac{\xbf_i \ybf_i}{\vert\vert  \xbf_i \vert \vert_2 \cdot \vert \vert \ybf_i \vert\vert_2})^2) $$,
	\end{array}
\end{equation}
where $\xbf_i$ and $\ybf_i$ represent the testing pixel corresponding to the scene at time $T_1$ and $T_2$, respectively. It is worth noting that the weight $\Wbf$ here is a monotonic increasing function, which is designed to further amplify the subtle differences between different signals, so as to further improve the separability.

\subsection{Fusion for Detection}

In this paper, the detection map based on morphological features is different from one after tensor reconstruction. They process original hyperspectral dataset in two different dimensions, spatial dimension and spectral dimension, respectively. However, for changed scenes, not every method can detect changed objects accurately. If one of the results fails to detect changed objects, it has a huge impact for detection accuracy. Therefore, fusion strategy can be adopted to solve this problem. Comparatively, average pooling operation is employed to avoid this issue, which is described as, 
\begin{equation}\label{eq18}
	\begin{array}{*{20}{l}}
		$$ R = a*R_1+b*R_2 $$,
	\end{array}
\end{equation}
where $R$ is the finally change detection result. $R_1$ and $R_2$ are these change maps corresponding to morphological domain and tensor domain, respectively.
In hyperspectral dataset, spatial information and spectral information are considered to be euqally important, so parameters $a$ and $b$ is set to the same weighted, that is, $a=b=\frac{1}{2}$.

\section{Experiments Results and Disscussion}
\label{sec3}

In following section, to validate the proposed dual-pipeline JMPT method effectively, two commonly used bitemporal datasets acquired by the Hyperion sensor are conducted to perform hyperspectral change detection. Detection results are compared with six other methods, including absolute distance (AD) \cite{du2012fusion}, Euclidian distance (ED) \cite{zhou2016novel}, absolute average difference (AAD) \cite{du2012fusion}, subspace-based change detector (SCD)\cite{wu2013subspace}, spectral angle weighted-based local absolute distance (SALA)\cite{hou2021hyperspectral}, and TDRD \cite{hou2021three}.


\subsection{Datasets Description}

\begin{table*}[tp]
	\caption{Thresholds setting under different attributes in max-tree/min-tree.}
	{ \scriptsize
		\begin{center}
			\begin{tabular}{cccccc}
				\hline
				{Attributes}  & {Area} & {Height} & {Volume} & {Diag\_box}  & {Std}\\
				\hline	
				Thresholds & 10,15,20,25,30,& 10,13,16,19,22,& 10,13,16,19,22,& 10,13,16,19,22,& 10,13,16,19,22,\\	
				~ &35,40,45,50,55&25,28,31,34,37&25,28,31,34,37& 25,28,31,34,37&25,28,31,34,37\\
				\hline
			\end{tabular}
			\label{tab:Thresholds}
	\end{center}}
\end{table*}

\begin{figure}[tp]
	\subfigure[\scriptsize{}]{
		\label{datasets_hermiston_t1}
		\includegraphics[width=1 in]{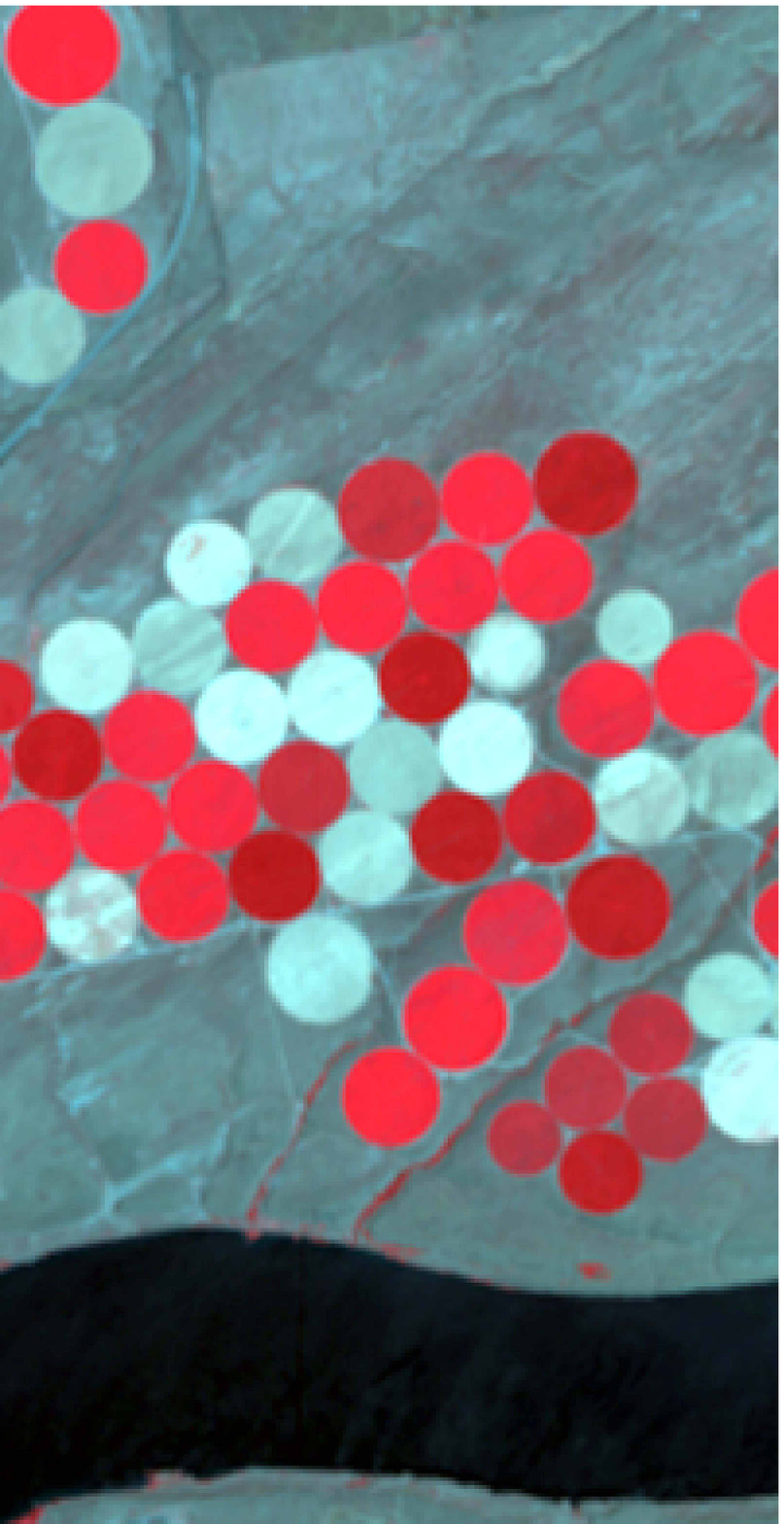}}
	\subfigure[\scriptsize{}]{
		\label{datasets_hermiston_t2}
		\includegraphics[width=1 in]{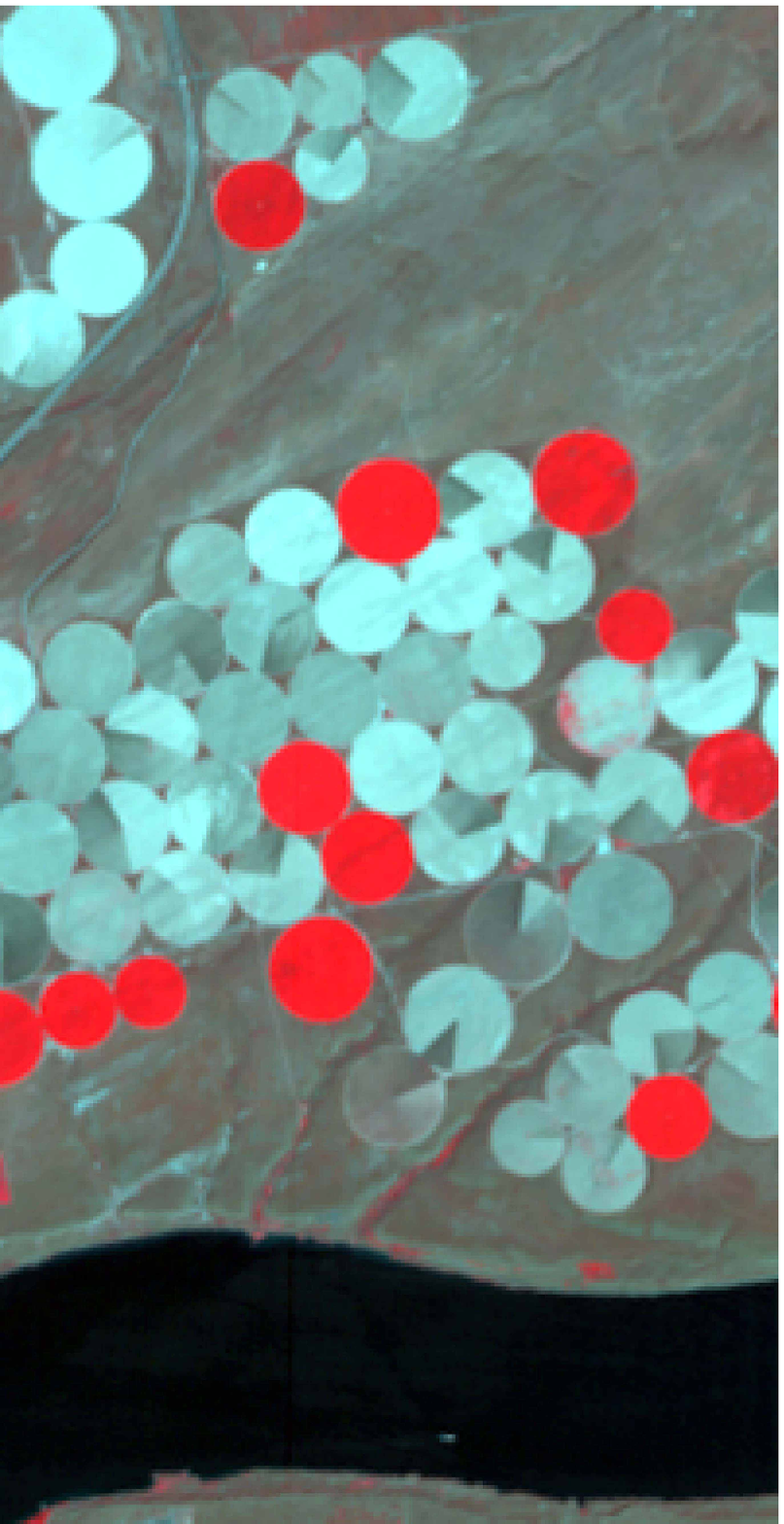}}
	\subfigure[\scriptsize{}]{
		\label{datasets_hermiston_gt}
		\includegraphics[width=1 in]{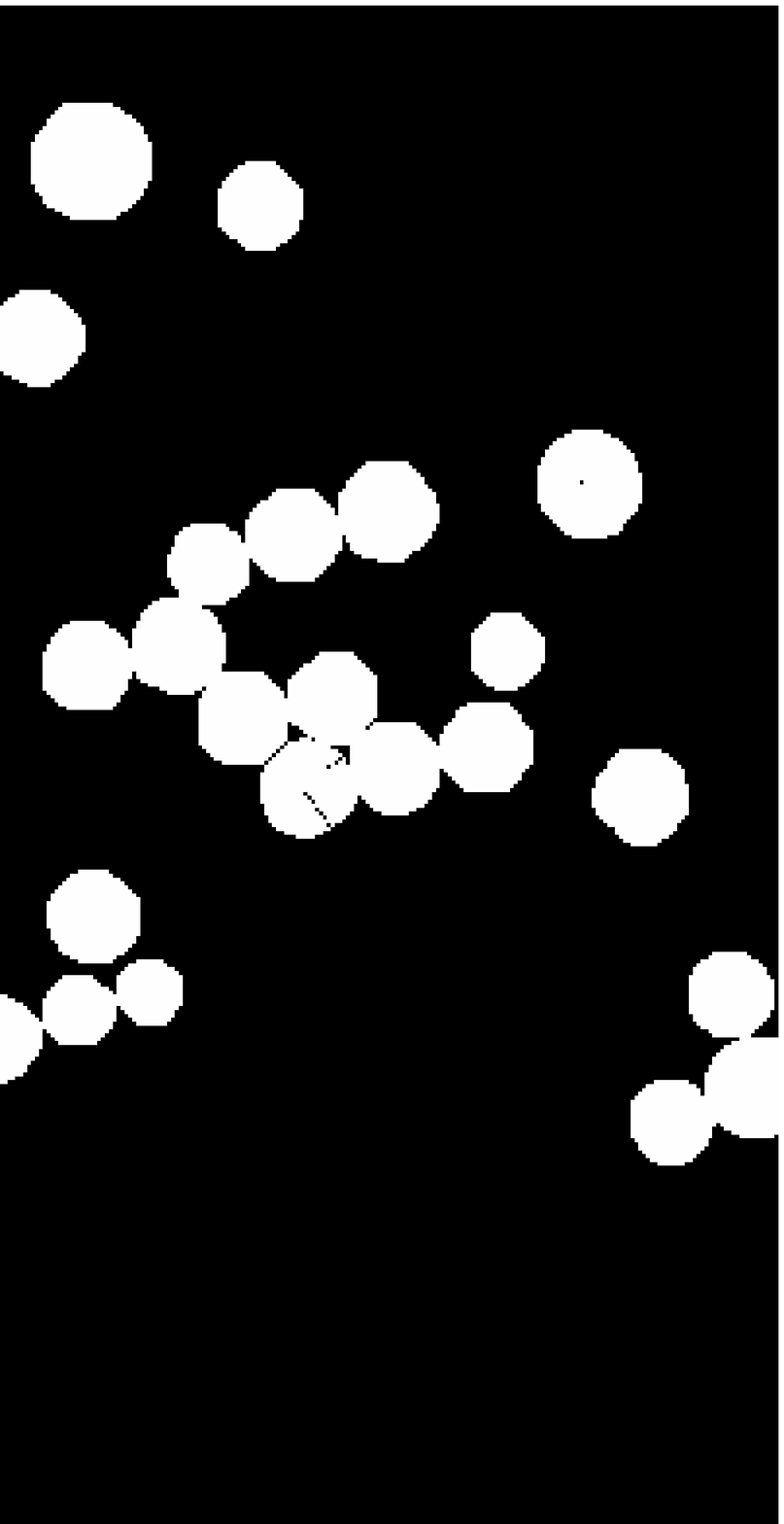}}
	
	\caption{Illustration of the Hermiston dataset. (a) The farmland on May 1, 2004. (b) The farmland on May 8, 2007. (c) The ground truth change map.}
	\label{Dataset_Hermiston}	
\end{figure}

\begin{figure}[tp]
	\subfigure[\scriptsize{}]{
		\label{datasets_yancheng_t1}
		\includegraphics[height=2.2 in, width=1 in]{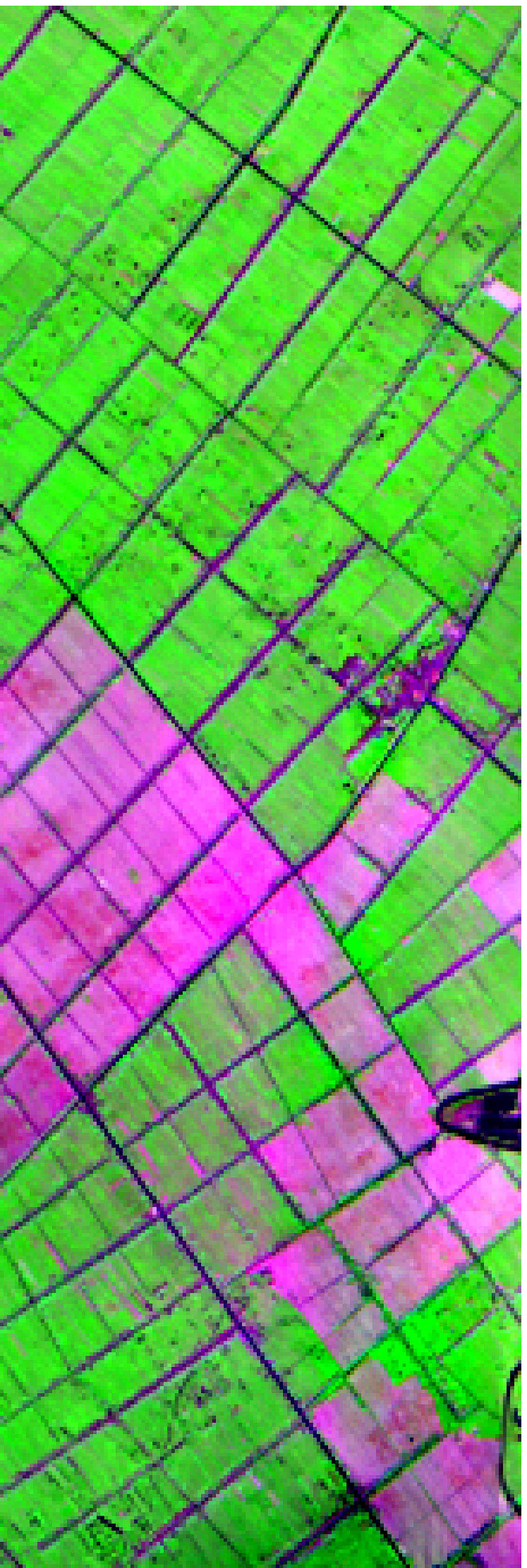}}
	\subfigure[\scriptsize{}]{
		\label{datasets_yancheng_t2}
		\includegraphics[height=2.2 in, width=1 in]{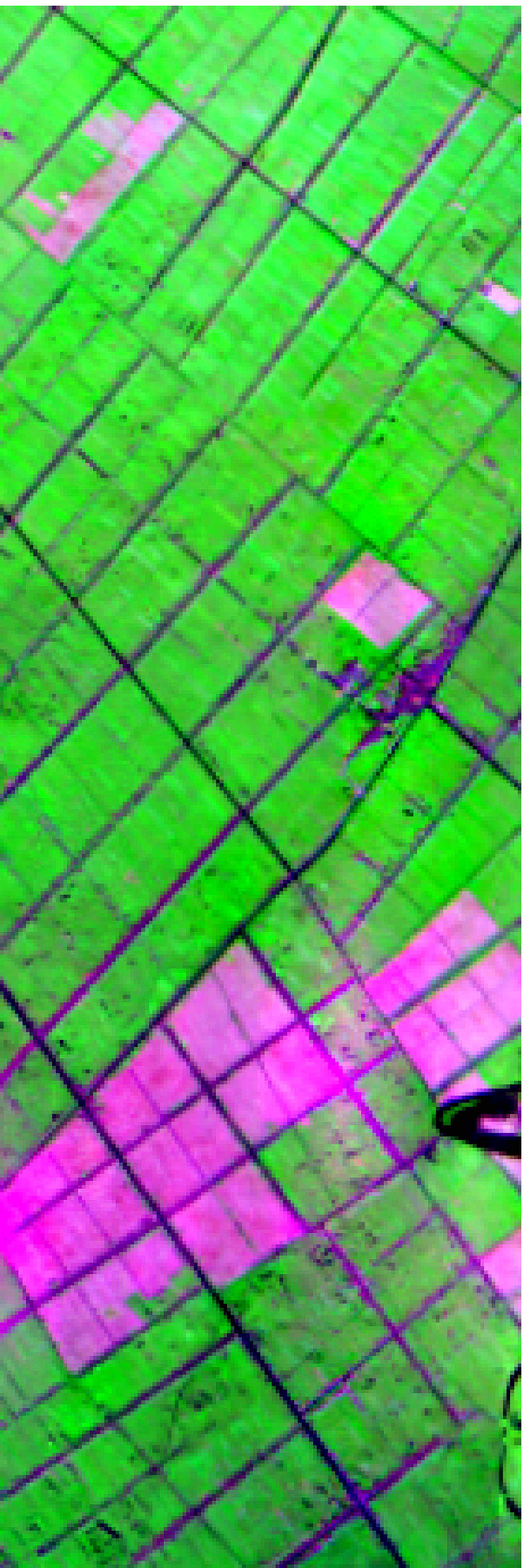}}
	\subfigure[\scriptsize{}]{
		\label{datasets_yancheng_gt}
		\includegraphics[height=2.2 in, width=1 in]{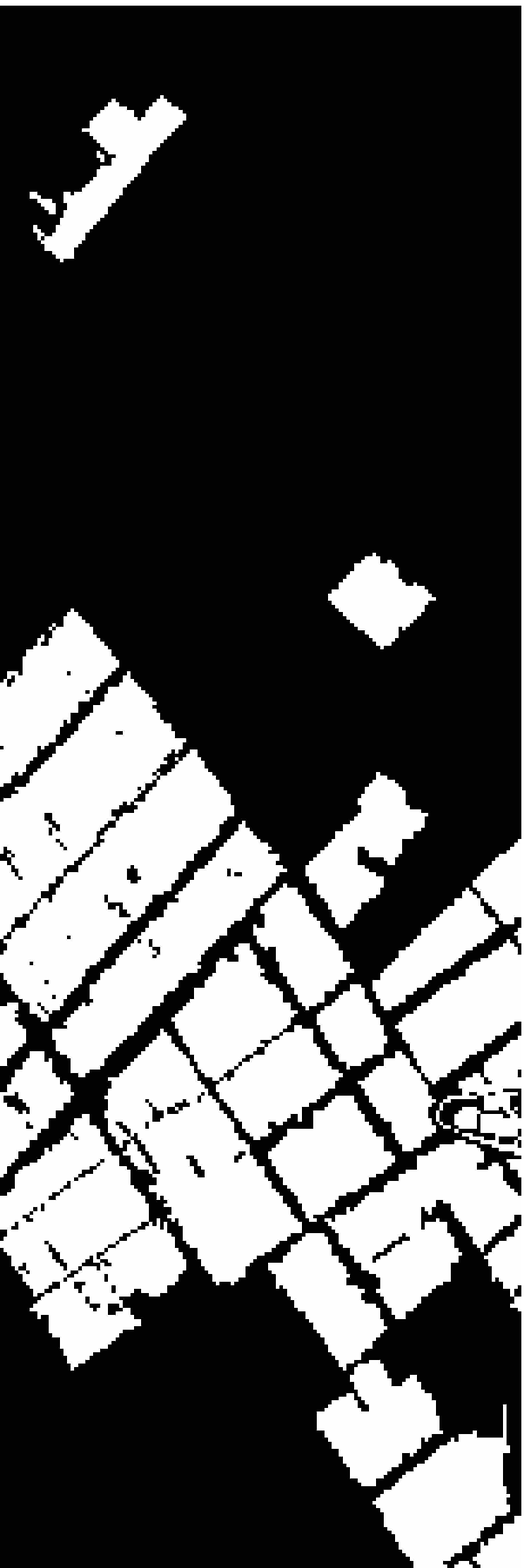}}
	
	\caption{Illustration of the Yancheng dataset. (a) The farmland on May 3, 2006. (b) The farmland on April 23, 2007. (c) The ground truth change map.}
	\label{Dataset_Yancheng}	
\end{figure}

The first dataset \cite{lopez2018stacked} is made of a pair of bi-temporal hyperspectral images, collected from an irrigated agricultural field in Hermiston City in Umatilla County, Oregon, United States, which were acquired on May 1, 2004 and May 8, 2007, respectively. It consists of 242 spectral bands and have the size of 390 $\times$ 200 pixels with the spatial resolution of 30 meter. The main change is farmland land-cover change, including the transitions among crops, soil, water and other land-cover types. The scene and the ground-truth map are shown in Fig. \ref{Dataset_Hermiston}. The second dataset including two bi-temporal images were collected from a wetland agricultural area in Yancheng city, Jiangsu Province, China on May 3, 2006 and April 23, 2007, respectively. The scene with 400 $\times$ 145 pixels and 154 spectral bands were used after removing noisy and water absorption bands. The main change type is also farmland land-cover change. The detailed images and the ground-truth map are illustrated in Fig. \ref{Dataset_Yancheng}

\subsection{Threshold Setting}

For different detection algorithms, the selected initial parameters is very important. Therefore, for fair comparison, an optimal parameter need to be selected for comparison. In TDRD method, it's detection performance is sensitive to parameter $\eta$, which is set empirically. According to reference\cite{hou2021three}, $\eta$ is set to $0.990$ and $0.955$ in Hermiston dataset and Yancheng dataset, respectively. For the proposed JMPT method, after constructing the max-tree and mini-tree, a suitable threshold value should be considered to prune the shape of the constructed tree, which is the filtering operation in the morphological processing. Among the five attributes, 10 thresholds are set for each attribute, and the minimum threshold is set 10. Interval of different thresholds in the area attribute is 5, and interval of other attributes is 3. It is worth noting that these parameters are set empirically. In current morphological-based algorithms, there is no effective automatic threshold selection strategy. Most of the thresholds are scene-specific. In this paper, because the threshold interval of area attribute is too small, the difference between various features is blurred, so here the interval of different thresholds is set as 5, and the interval of other four parameters is set as 3. The parameter setting is listed in Table \ref{tab:Thresholds}.

\begin{figure}[!t]
	\centering
	\begin{minipage}[c]{0.4\textwidth}
		\centering
		\includegraphics[width=2.8 in]{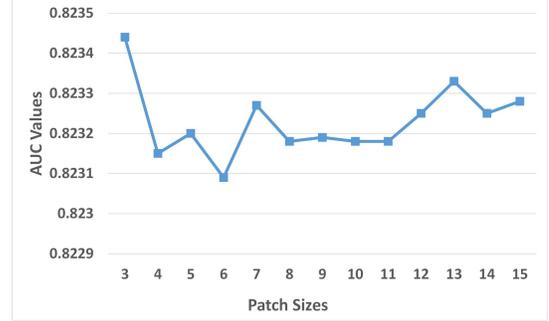}
		\centerline{(a)}		
	\end{minipage}
	\hspace{0.02\textwidth}
	\begin{minipage}[c]{0.4\textwidth}
		\centering
		\includegraphics[width=2.8 in]{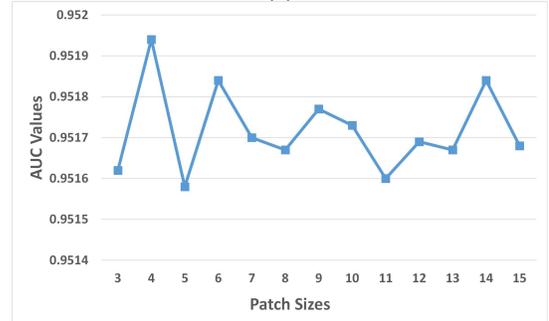}
		\centerline{(b)}
	\end{minipage}
	\caption{AUC values of different patch sizes: (a) Hermiston dataset, (b) Yancheng dataset}
	\label{patch_size}
\end{figure}

In tensor processing of JMPT,  change detection result is slightly sensitive to patch size $w$. If $w$ is too large or too small, it causes incomplete segmentation of image. Therefore, in order to ensure that the two bi-temporal images are as evenly divided as possible, $w$ is changed from 3 to 15, and detection maps (after tensor processing without morphological features) under different window sizes are also collected for evaluation\cite{liu2021multipixel}, where the optimal patch size is selected. These corresponding performances are shown in Fig. \ref{patch_size}, where we can observe that when patch size $w$ is set to $3$ and $4$ in Hermiston dataset and Yancheng dataset, respectively, the optimal detection accuracy is obtained. Therefore, $3$ and $4$ are selected as the optimal parameters for further comparison.

\subsection{Results and Discussion}

\begin{figure}[!t]
	\footnotesize
	\centering
	\begin{minipage}[c]{0.4\textwidth}
		\centering
		\includegraphics[width=2.8 in]{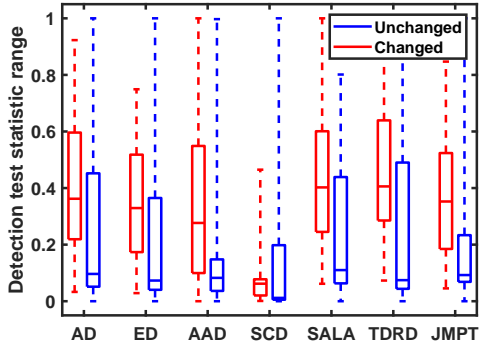}
		\centerline{(a)}		
	\end{minipage}
	\hspace{0.02\textwidth}
	\begin{minipage}[c]{0.4\textwidth}
		\centering
		\includegraphics[width=2.8 in]{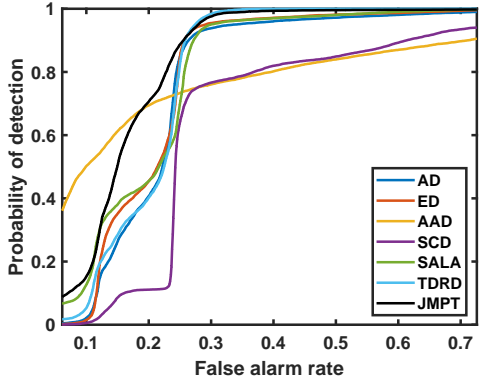}
		\centerline{(b)}
	\end{minipage}
	\caption{ROC curves of different methods in the Hermiston dataset. (a) Statistical separability analysis. (b) ROC curves.}
	\label{Hermiston_Results}
\end{figure}

\begin{figure}[!t]
	\footnotesize
	\centering
	\begin{minipage}[c]{0.4\textwidth}
		\centering
		\includegraphics[width=2.8 in]{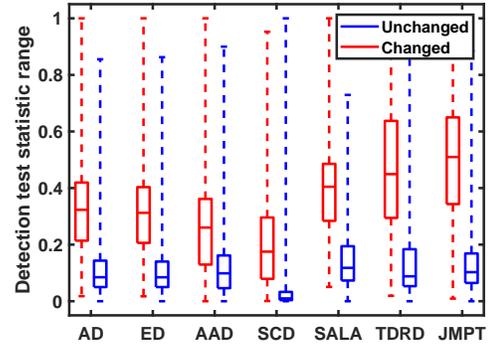}
		\centerline{(a)}		
	\end{minipage}
	\hspace{0.02\textwidth}
	\begin{minipage}[c]{0.4\textwidth}
		\centering
		\includegraphics[width=2.8 in]{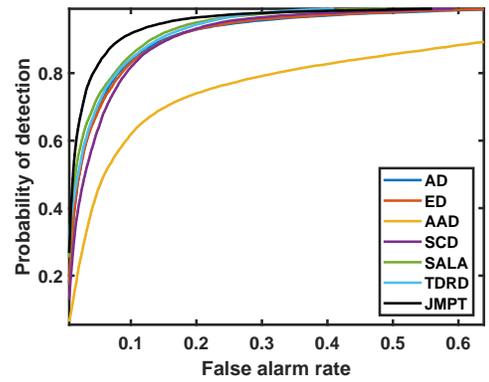}
		\centerline{(b)}
	\end{minipage}
	\caption{ROC curves of different methods in the Yancheng dataset. (a) Statistical separability analysis. (b) ROC curves.}
	\label{Yancheng_Results}
\end{figure}
\begin{table*}   
	\footnotesize
	\centering
	\caption{AUC values of various detectors using different datasets (AUC values \%).}
	\begin{tabular}{cccccccc}
		\hline
		Methods & AD & ED & AAD & SCD & SALA & TDRD & JMPT\\
		\hline
		Hermiston &78.582  &79.662  &77.828 &68.348 &80.445 &80.215 & \bf{83.769} \\
		Yancheng  &93.696  &93.667  &81.365 &93.443  &95.004 &95.275 & \bf{96.119} \\
		\hline
	\end{tabular}
	\label{Tab:AUC_Values}
\end{table*}   

\begin{table*}   
	\footnotesize
	\centering
	\caption{Execution time of various detectors using different datasets (unit: seconds).}
	\begin{tabular}{cccccccc}
		\hline
		Methods &AD &ED &AAD &SCD &SALA &TDRD &JMPT\\
		\hline
		Hermiston &0.040  &0.069  &0.043 &0.215  &4.120 &14.756 &190.62 \\
		Yancheng  &0.020  &0.040  &0.024 &0.150  &2.116 &7.014 &49.40 \\
		\hline
	\end{tabular}
	\label{Tab:Execution_Time}
\end{table*}   
Statistical separability analysis\cite{tan2019anomaly,liu2021multipixel}, also known as boxplot, is widely employed in hyperspectral anomaly detection, target detection, and change detection for performance assessment. In this paper, boxplot is used mainly to quantitatively compare distribution characteristics of multiple groups of data. In the statistical separability analysis, red box represents changed pixels and blue box represents unchanged pixels, where interval between the red box and the blue box represents separability between changed pixels and unchanged pixels. The upper and lower boundaries of these boxes are 80\% and 20\% of the statistical interval, while the 0\% - 20\% and 80\% - 100\% intervals are shown as dashed lines. The line in the box middle represents the median of dataset. The height of the blue box represents suppression degree of these methods to unchanged pixles. Generally, the lower blue box height is, the stronger the background suppression is, which is also conductive to separating changed pixels from unchanged ones. As shown in Fig. \ref{Hermiston_Results}(a), the separability of Hermiston dataset is displayed, where these methods show a relatively inferior ability to separate changed pixels from the background. However, after careful comparison, some subtle differences can also be found that the interval between red and blue boxes of JMPT are slightly larger than others, which shows that the separation ability of JMPT is slightly better than others. Similarly, for the Yancheng dataset shown in Fig. \ref{Yancheng_Results}(a), interval between red and blue boxes of JMPT is obviously larger than other methods, which indicates that the JMPT can separate changed pixels from unchangd pixels more effectively.

For more accurately compare the performance of various detection methods, receiver operating characteristic (ROC) \cite{hanley1982meaning} and area under the curve (AUC) metric are utilized as main criteria for evaluation. In ROC curve graph, the closer to the upper left corner, the larger the AUC, indicating that performance of the method is stronger. In Figs. \ref{Hermiston_Results}(b) and \ref{Yancheng_Results}(b), ROC curves of various detectors are illustrated. It is easy to find that the black curve representing JMPT is obviously on the upper left corner, which relects that the performance of JMPT is better than other methods. 

Table \ref{Tab:AUC_Values} and Table \ref{Tab:Execution_Time} provide AUC values and execution times of various detection methods, respectively. By analyzing, it is further confirmed that JMPT has better detection performance than other methods, which shows that collaborative processing of joint morphology and tensor can effectively improve the ability of change detection. All the experiments are conducted on windows 10 with 64-bit operating system. The processor of the system is Intel Core i7-8700 CPU with 3.20 GHz and 16 GB memory. From Table \ref{Tab:Execution_Time}, conclusion can be drawn that the computational cost of JMPT is high, especially for Hermiston dataset. The reason is that morphological feature extraction, and Tucker decomposition of patch-tensor are time-consuming.

\section{Conclusions}
\label{sec4}
In this paper, a novel dual-pipeline framework was designed for hyperspectral change detection, where morphological attribute profiles and tensor processing were effectively combined for hyperspectral change detection for the first time. Meanwhile, to the best of our knowledge, the patch-tensor processing were also used for hyperspectral change detection for the first time. Finally, a new detector was designed for further enlarge the difference between local pixels and testing pixel. Experimental conducted on two reals datasets demonstrated that the proposed detector achieved better detection performance. However, the computational cost was time-consuming, which is also the focus of our future work.

\bibliographystyle{IEEEtran}
\bibliography{bib_hzf}
\end{document}

%% file: com.tex
\newcommand{\ba}{\begin{array}}
\newcommand{\ea}{\end{array}}
\newcommand{\be}{\begin{displaymath}}
\newcommand{\ee}{\end{displaymath}}
\newcommand{\ben}{\begin{equation}}
\newcommand{\een}{\end{equation}}
\newcommand{\bena}{\begin{eqnarray}}
\newcommand{\eena}{\end{eqnarray}}
\newcommand{\beqa}{\begin{eqnarray*}}
\newcommand{\enqa}{\end{eqnarray*}}

\newcommand{\bc}{\begin{center}}
\newcommand{\ec}{\end{center}}
\newcommand{\bi}{\begin{itemize}}
\newcommand{\ei}{\end{itemize}}
\newcommand{\benu}{\begin{enumerate}}
\newcommand{\eenu}{\end{enumerate}}
\newcommand{\bdes}{\begin{description}}
\newcommand{\edes}{\end{description}}
\newcommand{\bt}{\begin{tabular}}
\newcommand{\et}{\end{tabular}}

\newcommand \xbf{{\bf x}}
\newcommand \ybf{{\bf y}}

\newcommand \Fbf{{\bf F}}

\newcommand \Ibf{{\bf I}}

\newcommand \Wbf{{\bf W}}

\newcommand{\circlambda}{\mbox{$\Lambda$
             \kern-.85em\raise1.5ex
             \hbox{$\scriptstyle{\circ}$}}\,}

